\documentclass[sigconf]{acmart}
\AtBeginDocument{%
  }

\copyrightyear{2026}
\acmYear{2026}
\setcopyright{cc}
\setcctype{by-nc-nd}
\acmConference[CIKM '26]{Proceedings of the 35th ACM International Conference on Information and Knowledge Management}{November 07--11, 2026}{Rome, Italy}
\acmBooktitle{Proceedings of the 35th ACM International Conference on Information and Knowledge Management (CIKM '26), November 07--11, 2026, Rome, Italy}
\acmDOI{10.1145/3799682.3840883}
\acmISBN{979-8-4007-2539-5/2026/11}

\usepackage[dvipsnames]{xcolor}
\usepackage[table]{xcolor}
\usepackage{color, colortbl}

\usepackage{multirow}
\usepackage{graphicx}

\usepackage{makecell}

\usepackage{amsmath}
\DeclareMathOperator*{\argmin}{arg\,min}

\begin{document}

\title{Neither Black nor White: Balancing Semantic and Collaborative Signals with Graph-Informed Semantic IDs (GrIS)}

\author{Aleksei Medvedev}
\authornote{These authors contributed equally to this work.}
\email{aleksei.medvedev@huawei.com}
\affiliation{%
  \institution{Huawei Ireland Research Centre}
  \city{Dublin}
  \country{Ireland}
}

\author{Alejandro Ariza-Casabona}
\authornotemark[1]
\email{alejandro.ariza.casabona@h-partners.com}
\affiliation{%
  \institution{Huawei Ireland Research Centre}
  \city{Dublin}
  \country{Ireland}
}

\author{Steven Derby}
\email{steven.derbi@huawei-partners.com}
\affiliation{%
  \institution{Huawei Ireland Research Centre}
  \city{Dublin}
  \country{Ireland}
}

\author{Gonzalo Fiz Pontiveros}
\email{gonzalo.fiz.pontiveros@huawei.com}
\affiliation{%
  \institution{Huawei Ireland Research Centre}
  \city{Dublin}
  \country{Ireland}
}

\author{Xinyang Shao}
\email{xinyang.shao@huawei-partners.com}
\affiliation{%
  \institution{Huawei Ireland Research Centre}
  \city{Dublin}
  \country{Ireland}
}

\author{Florian Spiess}
\email{florian.spiess@h-partners.com}
\affiliation{%
  \institution{Huawei Ireland Research Centre}
  \city{Dublin}
  \country{Ireland}
}

\renewcommand{\shortauthors}{Aleksei Medvedev et al.}

\begin{abstract}
Existing work on Semantic IDs (SIDs) for generative recommendation treats SID construction as a representation learning problem: encode items into a quantised latent space and read off codes. We argue this view is incidental.
SID construction is, at heart, a recursive clustering problem, and once stated this way the natural object to cluster is a graph whose nodes carry semantic content and whose edges carry collaborative signal; SID assignment becomes a hierarchical graph partition.
This reframing yields a unified framework, Graph-Informed Semantic IDs (GrIS), that subsumes prior approaches rather than displacing them.
RQ-VAE and RQ-KMeans are recovered as the special case where the graph is empty, exposing content-only quantisation as one corner of a larger design space along two so-far-collapsed axes: graph construction and recursive partition algorithm.
We explore two contrasting instantiations: RecDMoN, which performs hierarchical assignment via differentiable graph pooling, and RQ-GAE, which extends RQ-VAE with graph-aware item representations and a graph reconstruction objective.
On multiple real-world datasets, GrIS consistently improves over CF-aware SOTA, with gains of up to +52\% Hit@10. Because graph construction and partition are explicit, separately configurable components, improvements on either axis can be combined and evaluated systematically.

\end{abstract}




\maketitle

\section{Introduction}

Recommender systems are increasingly shifting towards generative prediction, where a model directly produces the identifier of the next item rather than retrieving candidates by nearest-neighbour search over dense embeddings. Central to this approach are Semantic IDs (SIDs): short sequences of discrete codes assigned to each item and generated autoregressively by the recommender. Prior work has shown that SIDs can improve efficiency, generalisation, and retrieval quality ~\cite{rajputRecommenderSystemsGenerative2023}, attracting significant research interest in how to construct better item tokenisations. Recent industrial studies further report online gains from SID-based retrieval at scale \cite{deng2025onerec,fu2026forge,yin2026dos,damico2026glide,ju2026snapchat}.

Most existing approaches formulate SID construction as a representation learning problem: encode each item into a dense latent vector, then quantise that vector into a sequence of codes. However, this framing is incomplete. An SID is not merely a compact representation of an item; it should organise items hierarchically from coarse to fine granularity, collapsing similar items into shared codes while preserving distinctions that matter for recommendation. These are properties of how the item space is structured and recursively partitioned for recommendation, and are not necessarily present in a good latent representation.

The gap becomes critical once collaborative signal, which is key for recommendation, is taken into account.
Supported by recent studies~\cite{wangLETTER2024, wangEAGER2024}, content similarity and behavioural similarity do not always coincide: semantically similar items can serve different audiences, while items frequently consumed together may share little surface content.
As a result, naive quantisation based solely on content embeddings leads to semantically plausible yet misaligned SID codes with the downstream recommendation objective.
Existing approaches respond by incorporating user-item graphs~\cite{liuMMGRecMultimodalGenerative2024}, collaboratively trained embeddings~\cite{wangLETTER2024}, or contrastive alignment into the SID pipeline~\cite{hePLUM2026}, proving that collaborative information matters for code assignment.
Nonetheless, these additions are typically introduced as modifications to representation learning, rather than as evidence that the primary objective of SID construction should itself be relational.

We argue that SID construction is, at heart, a recursive clustering problem and that the natural object to cluster is not an isolated item embedding but an item graph whose nodes carry semantic content and whose edges carry collaborative signal, leading to a hierarchical graph partition problem.
This perspective yields a unified framework, \textbf{Graph-Informed Semantic IDs (GrIS)}, that clarifies the structure of the design space and places several existing methods within it.
In particular, powerful residual quantisation baselines such as RQ-VAE~\cite{leeAutoregressiveImageGeneration2022} and RQ-KMeans~\cite{luoQARMQuantitativeAlignment2025} are recovered as the special case in which the graph is empty, i.e. where partitioning is driven only by node content and no relational signal is present.

The framework makes explicit two design choices that were previously treated as a single step: how to construct the graph and how to recursively partition it.
This separation makes it possible to reason about SID quality beyond a single tokeniser architecture and it turns seemingly disparate methods into comparable instances of the same principle.
Under GrIS, stronger semantic features can be combined with richer collaborative graphs and more effective hierarchical partitioners without reformulating the SID problem each time.

To demonstrate this, we study two contrasting instantiations of GrIS.
The first, \textbf{RecDMoN}, performs SID assignment through recursive differentiable graph pooling, directly optimising hierarchical partitions over the item-item graph.
The second, \textbf{RQ-GAE}, extends residual quantisation with graph-aware item representations and a graph reconstruction objective, showing how graph information can be injected into a quantisation-based pipeline without abandoning its basic structure.
Together, these instantiations illustrate that GrIS is not a single algorithm but a framework that spans multiple families of SID constructors.

Our contributions are fourfold:

\begin{itemize}
    \item We introduce GrIS, a new formulation of SID construction as hierarchical graph partition rather than stand-alone graph quantisation.
    \item We show that prior content-only SID methods arise as special cases of this formulation when collaborative edges are omitted.
    \item We explore two substantially different instantiations of our framework (RecDMoN and RQ-GAE), thereby exposing its own flexibility.
    \item Experimental results on multiple real-world datasets show that graph-informed SID construction yields consistent gains over strong collaborative and SID-based baselines, with improvements of up to 52\% Hit@10.
\end{itemize}

The rest of the paper is organised as follows: Section~\ref{sec:rel_work} reviews content-based, behaviour-enhanced, and structure-aware Semantic IDs in recent literature.
Section~\ref{sec:preliminaries} introduces the problem formulation and important notation.
Section~\ref{sec:framework} formalises GrIS, shows how existing methods fit into the framework, and instantiates RecDMoN and RQ-GAE.
Section \ref{sec:experiments} reports the experimental setup, results, and discuss implications, and limitations of each approach.
Finally, Section~\ref{sec:conclusion} presents conclusions and directions for future work.

\section{Related Work}\label{sec:rel_work}

Existing work on semantic identifiers for generative recommendation mainly differs in which signals are encoded into the tokenisation space. We organise the literature into content-based, behaviour-enhanced, and structure-aware semantic IDs. We additionally review two classes of graph-processing operations that are central to our framework: graph convolution and graph pooling.

\subsection{Content-based Semantic IDs}

Early semantic ID methods derive item tokens primarily from item content. TIGER~\cite{rajputRecommenderSystemsGenerative2023} is the representative example: it encodes item text into dense embeddings and uses RQ-VAE to quantise them into hierarchical code sequences. This approach showed that semantically meaningful IDs generalise better than random hashing or atomic IDs, especially for cold-start and semantic sharing across similar items.
More broadly, this family of methods treats item tokenisation as semantic quantisation over content features. Related quantisation schemes such as hierarchical clustering, product quantisation variants, and residual clustering methods \cite{houGeneratingLongSemantic2025,luoQARMQuantitativeAlignment2025,liuMBGen2024,fuDifferentiableSemanticID2026} follow the same principle: build compact discrete codes from content representations before training the generative recommender. Their common limitation is that the resulting IDs are governed by semantic similarity, which may not always align with user behaviour.

\subsection{Behaviour-enhanced Semantic IDs}

A second line of work responds by injecting collaborative signals into semantic ID learning. LETTER~\cite{wangLETTER2024} is a representative method in this direction. It keeps hierarchical semantic quantisation via RQ-VAE, but further aligns quantised representations with collaborative filtering embeddings, thus relying on a pretrained CF-based model, and adds diversity regularisation to reduce code assignment bias. PLUM~\cite{hePLUM2026} extends this idea in a large-scale Large Language Model (LLM) setting by combining multi-embedding content fusion with co-occurrence-based contrastive regularisation, so that user behaviour shapes the SID space during tokenisation rather than only during downstream training. Overall, these methods shift the goal from pure semantic compression to semantic-behaviour alignment.

\subsection{Structure-aware Semantic IDs}

A third line explicitly exploits structural information, such as co-occurrence graphs or interaction topology, during ID construction. In this setting, the structure itself helps define the semantic space instead of serving only as auxiliary supervision.

MMGRec~\cite{liuMMGRecMultimodalGenerative2024} is a representative graph-enhanced tokeniser: it uses Graph-aware RQ-VAE to fuse multimodal item features with collaborative signals from a user-item graph before quantisation, and further handles collisions with a popularity-based token. S2GR~\cite{guoS2GR2026} also belongs to this line, as it improves the SID space with item co-occurrence structure, codebook balancing, and hierarchical reasoning over the resulting IDs. In addition, \cite{huaHowIndexItem2023} is particularly relevant here because it explores several pure structure-based alternatives: identifiers such as CID or SID are derived from item co-appearance structure rather than content-based representations. This makes it a useful reference point for the structure-aware setting, especially when contrasting structure-first tokenisation against content-driven SID learning.

GrIS unifies all three lines by formalising SID construction as hierarchical graph partition, with content-only methods arising as the degenerate case of an empty graph.

\subsection{Graph Convolution and Graph Pooling}

Graph neural networks (GNNs) commonly process graph-structured data through neighbourhood aggregation. In graph-convolutional architectures, each node updates its representation using its own features and information propagated from adjacent nodes. Stacking multiple such layers allows information to be integrated from multi-hop neighbourhoods. These architectures have been successfully applied to link prediction~\cite{zhangLinkPredictionBased2018}, node classification~\cite{kipf2017semi,hamiltonInductiveRepresentationLearning2017}, and graph classification~\cite{xu2018how}. Several structure-aware semantic ID methods discussed above use this mechanism to enrich content-based item representations with relational signals from the interaction graph.

Graph pooling~\cite{yingHierarchicalGraphRepresentation2018} progressively coarsens a graph by learning soft node-to-cluster assignments and aggregating the corresponding features and connectivity. Beyond producing graph-level representations, this formulation provides a differentiable approach to discovering coherent node clusters without ground-truth assignments. MinCutPool~\cite{bianchiSpectralClusteringGraph2020} and DMoN~\cite{tsitsulinGraphClusteringGraph2023} learn such assignments jointly from graph topology and node attributes by optimising objectives based on normalised min-cut and modularity, respectively, together with regularisation terms that prevent degenerate partitions. Applying these operations recursively or across multiple stages yields hierarchies of graph-aware clusters, providing a natural basis for constructing multi-level semantic IDs.

\section{Preliminaries}\label{sec:preliminaries}

In this Section, we introduce the notation used throughout the paper and formalise the Semantic ID construction problem considered in generative recommendation.
\paragraph{\textbf{Generative Recommendation Setup}} Let $\mathcal{U} = {u_{1}, ..., u_{|\mathcal{U}|}}$ be the set of users and $\mathcal{I} = {i_{1}, ..., i_{|\mathcal{I}|}}$ the set of items. For each user $u \in \mathcal{U}$, we observe their interaction history $x_{u} = (i_{u, 1}, ..., i_{u, T_u})$, where $i_{u, t} \in \mathcal{I}$ denotes the item interacted with by the user at step $t$.
In sequential recommendation, the goal is to predict the next item $i_{u, T_u + 1}$ given the prefix $x_u$.

\begin{figure*}
    \centering
    \includegraphics[width=0.9\linewidth]{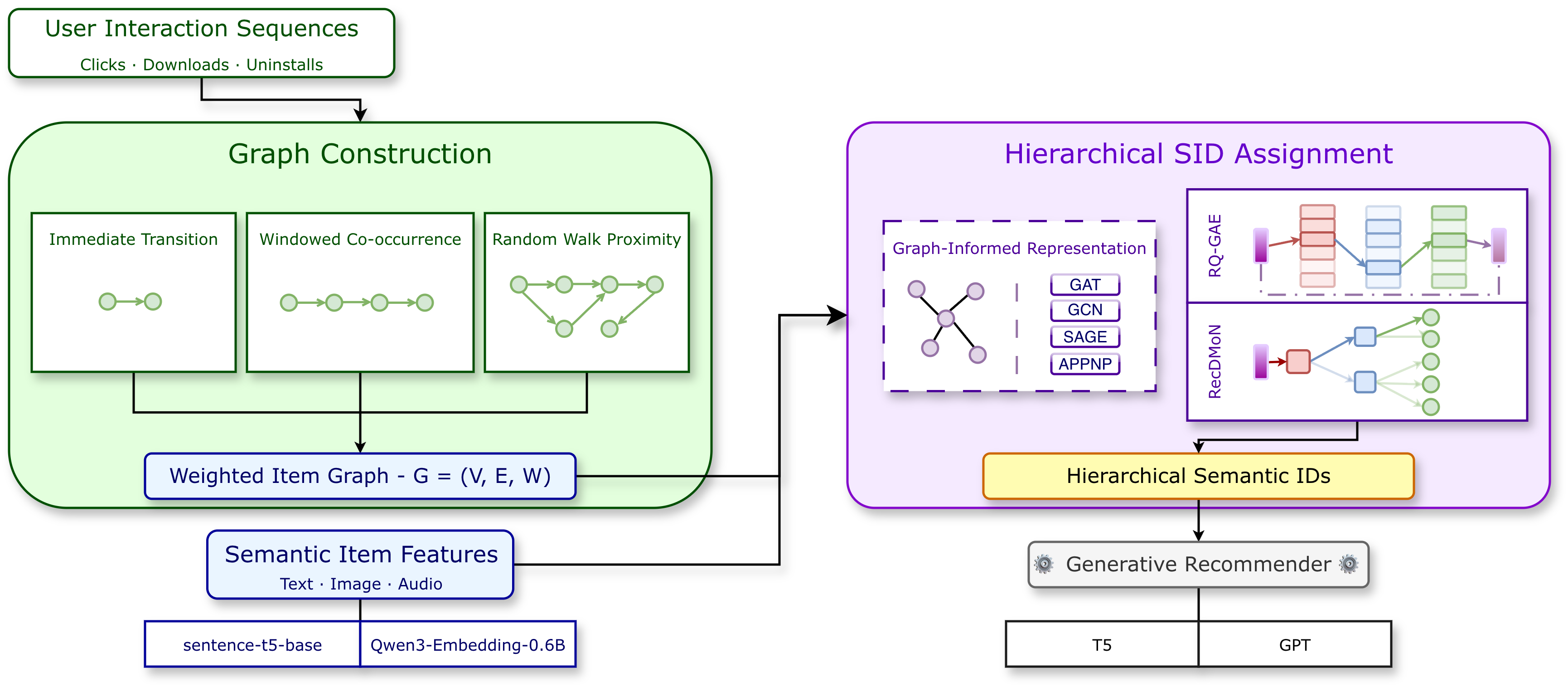}
    \caption{Overview of the GrIS framework with two main axes: Graph Construction and Hierarchical SID Assignment. Graph Construction uses the user behaviour data to construct a weighted item graph based on predefined construction rules. This graph together with semantic item representations are fed to the second module to perform hierarchical partition. The output hierarchical SIDs are used to train a Generative Recommender model. In particular, we explore RQ-GAE and RecDMoN instantiations of the GrIS framework.}
    \label{fig:framework}
\end{figure*}

\paragraph{\textbf{Item Semantic Representations}} Each item $i \in \mathcal{I}$ is associated with a set of semantic embeddings $\mathcal{E}_i = \{e_i^1, \ldots, e_i^{M_i}\}$, where each embedding is derived from content data associated with the item. While content data may span multiple modalities, in this work we use only text embeddings derived from item metadata, with the specific fields varying by dataset.

\paragraph{\textbf{Collaborative Graph}} User interaction sequences induce relational structure over items that can be 
represented as a weighted graph $G = (V , E, W)$ where $\mathcal{I}\subseteq V$, the edges $E \subseteq V \times V$ connect behaviourally related nodes, and $W : E \rightarrow \mathbb{R}_{>0}$ assigns weights that measure the strength of the relation. The specific construction rule that determines $E$ 
and $W$ from interaction data is a configurable component of the GrIS framework, 
discussed in Section~\ref{sec:framework}.

\paragraph{\textbf{Semantic IDs}} A Semantic ID (SID) is a map
\begin{align*}
    \mathrm{SID} &: \mathcal{I} \rightarrow \mathcal{C}^1 \times \cdots \times \mathcal{C}^L, \\
                 &\phantom{:}\ i \mapsto (c_i^1, \ldots, c_i^L),
\end{align*}
where $\mathcal{C}^1, \ldots, \mathcal{C}^L$ denote $L$ codebooks, one per level.\footnote{Recent work explores variable-length~\cite{khrylchenkoVariableLengthSemanticIDs2026}
and end-to-end~\cite{fuDifferentiableSemanticID2026} SID construction;
we leave integration with GrIS to future work.}

\paragraph{\textbf{Problem Statement}}

Given a set of items $\mathcal{I}$, their semantic embeddings 
$\{\mathcal{E}_i\}_{i \in \mathcal{I}}$, and user interaction sequences 
$\{x_u\}_{u \in \mathcal{U}}$, the goal is to construct a SID map 
whose hierarchical structure improves retrieval quality when 
used by a generative recommender. We focus exclusively on SID construction and hold the downstream generative recommender fixed.

\section{Graph Informed Semantic IDs (GrIS)}\label{sec:framework}

This Section introduces Graph Informed Semantic IDs (GrIS), a general framework for constructing hierarchical item identifiers from both semantic item features and collaborative item structure. Figure~\ref{fig:framework} illustrates the main components and potential design choices of our framework.

\subsection{Overview}\label{sec:overview}

GrIS formalises SID construction as hierarchical graph partition over an item graph whose nodes carry semantic content and whose edges carry collaborative signal. This formulation exposes two independent design axes that were previously treated as a single step: \textit{graph construction}, which determines what relational structure is encoded, and \textit{hierarchical partition}, which determines how that structure is recursively divided into codes. Each axis is an independently configurable component; Table~\ref{tab:instantiations} summarises how existing methods and our two instantiations occupy the resulting design space.

\begin{table*}
\centering
\caption{
Instantiations of the GrIS framework. Content-only methods arise as special cases with an empty graph ($E = \emptyset$).
}
\label{tab:instantiations}
\begin{tabular}{lll}
\toprule
\textbf{Method} & \textbf{Graph Construction} & \textbf{Hierarchical Partition} \\
\midrule
TIGER~\cite{rajputRecommenderSystemsGenerative2023} & $\emptyset$ & RQ-VAE \\
LETTER~\cite{wangLETTER2024} & $\emptyset$ &   RQ-VAE + Contrastive CF + Diversity Loss\\
\midrule
MMGRec~\cite{liuMMGRecMultimodalGenerative2024} & Multimodal Bipartite $\mathcal{U}\times\mathcal{I}$ & RQ-VAE + BPR loss \\
S$^2$GR~\cite{guoS2GR2026} & Windowed Co-occurrence $\mathcal{I}\times\mathcal{I}$ &   RQ-VAE + Load Balancing Loss\\
\midrule
RecDMoN (ours) & Immediate Transition $\mathcal{I}\times\mathcal{I}$ & Differentiable Graph Pooling (DMoN) \\
RQ-GAE (ours)  & Immediate Transition $\mathcal{I}\times\mathcal{I}$ & RQ-VAE + Graph Reconstruction Loss\\
\bottomrule
\end{tabular}
\end{table*}

\subsection{Item Graph Construction}
The first step in GrIS is to construct an item graph $G$ from interaction data. For simplicity, we focus on item-item graph construction, though other entities such as users, or external entities (Knowledge Graphs) could be considered as well. Given user sequences $\{x_u\}_{u\in\mathcal{U}}$, we define an edge between items $i$ and $j$ whenever they are considered related under a chosen graph construction rule, such as windowed co-occurrence, immediate sequential transition, session-based co-appearance, or random walk proximity over an interaction graph. A weight $\omega_{ij} = W(i, j)$ is assigned to each candidate edge based on the strength of the co-occurrence signal, and edges below a minimum threshold $\lambda$ may be discarded to control the graph's density . This abstraction gives flexibility to the framework to adapt to different recommendation settings: sequential edges emphasise temporal transitions, session graphs emphasise local co-consumption, and random walk-based edges capture broader multi-hop affinity.

\subsection{Hierarchical SID assignment}

Given a graph $G$ and initial item representations $X$, GrIS assigns each item a hierarchical SID.

\subsubsection{Graph-Informed Representations}

Once the graph $G$ is built, GrIS computes graph-informed item representations by propagating information across neighbours:
\begin{equation*}
    H = f_{GNN}(G, X)
\end{equation*}
where $X$ is the matrix of initial item embeddings and $h_i$ is the graph-informed representation of item $i$. This encoder may be parametric, such as a graph convolutional or attention-based network, or non-parametric, such as normalised neighbourhood aggregation with fixed propagation weights. In both cases, the aim is to enrich each item representation with local collaborative context from nearby items in the graph.

A generic K-layer propagation form can be written as
\begin{equation*}
    H^{k+1} = \phi(\hat{A} H^k W^k),\qquad H^0 = X
\end{equation*}
for parametric encoders where $W^k$ is learnable weight matrix, or as
\begin{equation*}
    H^{k+1} = \phi(\hat{A} H^k)
\end{equation*}
for non-parametric propagation, where $\hat{A}$ is a normalised adjacency operator and $\phi(\cdot)$ is a pointwise nonlinearity.

\subsubsection{Residual Quantisation Instantiation}

Starting from each graph-informed representation $h_i$, we map it into a latent vector $z_i = MLP(h_i)$ which is then discretised through residual quantisation across $L$ codebooks. Let the initial residual be $r_i^1 = z_i$. At level $l$, the code assignment is
\begin{equation*}
    c_i^l = \argmin_{k \in \mathcal{C}^l} \left|\left| r_i^l - b_k^l\right|\right|_2^2
\end{equation*}
where $b_k^l$ is the $k$-th code embedding in codebook $l$. The residual is updated recursively as
\begin{equation*}
    r_i^{l+1} = r_i^l - b_{c_i^l}^l.
\end{equation*}

\paragraph{Training Objective}

The residual quantisation objective follows the standard RQ-VAE-based SID construction. We augment it with a graph reconstruction term that encourages the quantised latent space to preserve the item--item interaction graph. Rather than reconstructing the full adjacency matrix, which is expensive for large item vocabularies, we optimise a subsampled neighbourhood softmax over mini-batches.

Let $q_i = \sum_{l=1}^{L} b_{c_i^l}^{l}$ denote the quantised latent representation of item $i$. Given a mini-batch $\mathcal{B}$, we treat items in $\mathcal{B}$ as sampled anchor and candidate nodes from the full item graph. Let $w_{ij}$ denote the edge weight between items $i$ and $j$ in the induced batch subgraph. We define the subsampled neighbourhood distribution as
\[
    p_\theta(j \mid i, \mathcal{B})
    =
    \frac{
        \exp(\bar{q}_i^\top \bar{q}_j / \tau)
    }{
        \sum_{k \in \mathcal{B},\, k \neq i}
        \exp(\bar{q}_i^\top \bar{q}_k / \tau)
    },
    \qquad
    \bar{q}_i = \frac{q_i}{\|q_i\|_2},
\]
where $\tau$ is a temperature parameter. The graph reconstruction loss is
\[
    \mathcal{L}_{\mathrm{graph}}
    =
    -
    \frac{1}{|\mathcal{B}_{+}|}
    \sum_{i \in \mathcal{B}_{+}}
    \sum_{j \in \mathcal{B}}
    w_{ij}
    \log p_\theta(j \mid i, \mathcal{B}),
\]
where
\[
    \mathcal{B}_{+}
    =
    \left\{
        i \in \mathcal{B}
        :
        \sum_{j \in \mathcal{B}} w_{ij} > 0
    \right\}.
\]
Thus, for each sampled anchor item, the model assigns higher probability to its observed graph neighbours within the batch. The remaining batch items form the normalising set of the subsampled softmax. This objective is related to weighted neighbourhood-likelihood objectives such as LINE~\cite{tang2015line}, but we apply it directly to the quantised latent representations used for SID construction.

The final objective is
\[
    \mathcal{L}
    =
    \mathcal{L}_{\mathrm{RQ}}
    +
    \beta \mathcal{L}_{\mathrm{graph}},
\]
where $\mathcal{L}_{\mathrm{RQ}}$ denotes the standard RQ-VAE reconstruction and quantisation objective, and $\beta$ controls the strength of graph-structural supervision.

\subsubsection{Recursive DMoN Instantiation}

This GrIS variant builds SIDs by recursively partitioning the item graph or the graph-informed latent space. Starting from the full item set, a clustering operator splits the current cluster into a small number of subclusters, and the branch selected for item $i$ at recursion depth $l$ becomes the $l$-th SID code $c_i^l$.

When Recursive DMoN is used, the hierarchy is induced by repeated graph clustering, so the SID no longer comes from codebook quantisation but from the path of the item in the partition tree. This approach is particularly interesting because it makes the graph structure explicit in the identifier itself. That is, two items share a prefix exactly when they remain in the same partition up to that recursion level.

\paragraph{Training Objective}
Following DMoN~\cite{tsitsulinGraphClusteringGraph2023}, each recursive split is trained as a local differentiable graph clustering problem.
Consider a parent cluster at recursion depth $l$ with induced weighted adjacency matrix $A \in \mathbb{R}^{n \times n}$ and node representations $X$.
A local assignment network produces soft assignments of the $n$ nodes to $K$ child clusters:
\begin{equation*}
    S = \operatorname{softmax}(g_\theta(X, A)) \in \mathbb{R}^{n \times K},
\end{equation*}
where $K$ is the fixed branching factor shared across all recursion levels.
Let $d = A\mathbf{1}$ be the degree sequence, $m = \frac{1}{2}\mathbf{1}^{\top}d$ the number of edges, and set
\begin{equation*}
    B = A - \frac{d d^\top}{2m},
\end{equation*}
which is known as the local modularity matrix.
For each parent cluster, the assignment network is optimised using the DMoN objective:
\begin{equation*}
    \mathcal{L}
    =
    -
    \frac{1}{2m}
    \operatorname{Tr}\!\left(S^\top B S\right)
    +
    \frac{\sqrt{K}}{n}
    \left\|
        \sum_{i=1}^{n} S_i
    \right\|_F
    -
    1 .
\end{equation*}
The first term maximises relaxed graph modularity, encouraging items with stronger-than-expected connectivity to remain in the same child partition.
The second term is the DMoN collapse regularisation term, which penalises highly imbalanced soft assignments across child clusters.
After optimising the local split, each item is assigned to the child cluster with the largest assignment probability:
\begin{equation*}
    c_i^l
    =
    \operatorname*{arg\,max}_{k \in \{1,\ldots,K\}}
    S_{ik}.
\end{equation*}
The procedure is then applied recursively to the resulting child clusters.
The SID of item $i$ is given by the sequence of branch indices along its root-to-leaf path.

\section{Experiments}\label{sec:experiments}

\subsection{Experimental Setup}

\paragraph{Datasets} We evaluate the proposed framework on multiple recommendation benchmarks spanning different domains: four subsets from the Amazon Reviews 2014 dataset \cite{heAmazon2016} (Beauty, Toys, Sports, and Books), MIND-Small \cite{wuMIND2020} (a public news recommendation dataset released by Microsoft), and Yelp 2018\footnote{https://www.yelp.com/dataset} (a user-business interaction dataset from the Yelp platform). These datasets provide a diverse test bed covering e-commerce, news, and local business scenarios, allowing us to assess the benefits and drawbacks of each instantiation of the framework across distinct item semantics, interaction patterns, sparsity levels, and catalogue sizes. Table \ref{tab:data_stats} shows the statistics of the selected datasets.

\begin{table}
    \centering
    \caption{Dataset statistics.}
    \label{tab:data_stats}
    \resizebox{\columnwidth}{!}{%
    \begin{tabular}{lrrrrr}
    \toprule
    Dataset & $|\mathcal{U}|$ & $|\mathcal{I}|$ & \#Interactions & Avg. $T_u$ & Sparsity \\
    \midrule
    Toys & 19,412 & 11,924 & 167,597 & 8.63 & 99.928\% \\
    Beauty & 22,363 & 12,101 & 198,502 & 8.88 & 99.927\% \\
    Sports & 35,598 & 18,357 & 296,337 & 8.32 & 99.955\% \\
    MIND & 86,913 & 20,283 & 2,308,143 & 26.56 & 99.869\% \\
    Yelp & 213,170 & 94,304 & 3,277,932 & 15.38 & 99.984\% \\
    Books & 603,668 & 367,982 & 8,898,041 & 14.74 & 99.996\% \\
    \bottomrule
    \end{tabular}}
\end{table}

\paragraph{Baselines}
We consider representative baselines from multiple methodological families:
\begin{itemize}
    \item[-] \textit{CF-only atomic ID methods:}
    \textbf{SASRec}~\cite{kangSASRec2018}, a self-attentive sequential recommender over atomic item IDs, and \textbf{HSTU}~\cite{zhaiHSTU2024}, a strong and scalable sequential recommender with relative position and temporal attention biases over the original user-item ID space.
    
    \item[-] \textit{Semantic-only SID generative method:}
    \textbf{TIGER}~\cite{rajputRecommenderSystemsGenerative2023}, which represents items using semantic IDs obtained from residual quantisation and performs recommendation via generative decoding over semantic tokens rather than conventional item ranking.
    
    \item[-] \textit{CF-only SID method:}
    \textbf{Spectral clustering}~\cite{huaHowIndexItem2023}, which constructs semantic-style item assignments purely from collaborative filtering structure.
    
    \item[-] \textit{Graph-free SID assignment method:}
    \textbf{LETTER}~\cite{wangLETTER2024} learns hierarchical SIDs
without an explicit item graph, injecting collaborative signals through
a pretrained SASRec auxiliary model and regularisation losses.
    
    \item[-] \textit{Graph-aware GrIS methods:}
    \textbf{MMGRec}~\cite{liuMMGRecMultimodalGenerative2024}, which integrates multimodal and graph signals within the GrIS framework, and \textbf{S$^2$GR}~\cite{guoS2GR2026}, which enhances hierarchical semantic ID learning by explicitly modelling structural relations in the graph.
\end{itemize}
\paragraph{Implementation Details}

Following prior work~\cite{houGeneratingLongSemantic2025,juGRID2025,wangLETTER2024}, we apply iterative 5-core filtering to all datasets and adopt a leave-one-out evaluation protocol, where each user's last interaction is used for testing and the second-to-last interaction is used for validation. Furthermore, we fix the maximum sequence length to 20 following~\cite{wangLETTER2024}.
We report Hit Rate (\textbf{H}) and normalised Discounted Cumulative Gain (\textbf{N}) at cutoffs 5 and 10, with their scores expressed as percentages.
Each experiment is repeated with three random seeds and average performance across seeds is reported.

For all residual-quantisation-based SID mapping methods, including RQ-GAE, we use four codebooks with 256 entries each, following~\cite{wangLETTER2024}.
SID mapping models are trained for 20K epochs using AdamW with learning rate $10^{-3}$, batch size 1024, and weight decay $10^{-4}$; for the Books dataset, we increase the batch size to 8192 due to its larger scale.
Downstream generative recommendation models are trained for 150 epochs with AdamW, learning rate $3 \times 10^{-4}$ and early stopping patience of 20 based on validation performance.

For RQ-GAE, we use APPNP~\cite{klicperaAPPNP2019} as the graph propagation operator and construct the item graph using adjacent co-occurrence edges from training sequences.
For S2GR, we omit the reasoning component and use the same downstream generative recommendation backbone as the other SID methods, since our study focuses on isolating the effect of SID construction rather than comparing full recommendation frameworks.
For MMGRec, we follow the public implementation and hyperparameter setting of~\cite{liuMMGRecMultimodalGenerative2024}: the GCN component is first trained for 1000 epochs with BPR loss, after which the RQ-VAE stage is trained using the same residual-quantisation setup as the other SID baselines and RQ-GAE.

Additional implementation details are provided in our source code at \url{https://github.com/hirc-airecs/graph-informed-sids}.

\subsection{Experimental Results}

Our experiments address the following questions: (a) Can GrIS instantiations improve over content-only and CF-aware SID baselines? (b) How does graph construction affect downstream performance? (c) How do different hierarchical SID assignment techniques behave when combined with graph-informed representations?

\subsubsection{Generative Recommendation Comparison}

\begin{table*}
\centering
\renewcommand{\arraystretch}{0.78}
\caption{Results across datasets. Bold scores indicate leading results while underscore indicates second best results. $^\dagger$ denotes significant improvement over the best baseline.}
\label{tab:results_split_by_dataset}
\begin{tabular}{l|cccc@{\hspace{5pt}}|cccc}
\toprule
\textbf{Method} & \textbf{H@5} & \textbf{H@10} & \textbf{N@5} & \textbf{N@10} & \textbf{H@5} & \textbf{H@10} & \textbf{N@5} & \textbf{N@10} \\
\midrule
& \multicolumn{4}{c|}{\textbf{Books}} & \multicolumn{4}{c}{\textbf{Beauty}} \\
\midrule
\textbf{RecDMoN} & -- & -- & -- & -- & \textbf{4.89}$^\dagger$ & \textbf{7.96}$^\dagger$ & \textbf{3.11}$^\dagger$ & \textbf{4.09}$^\dagger$ \\
\rowcolor{gray!20}$\Delta_{\mathrm{LETTER}}$ & -- & -- & -- & -- & +21.7\% & +28.9\% & +14.6\% & +20.2\%  \\
\textbf{RQ-GAE} & \textbf{3.31}$^\dagger$ & \textbf{4.60}$^\dagger$ & \textbf{2.39}$^\dagger$ & \textbf{2.81}$^\dagger$ & \underline{4.43}$^\dagger$ & \underline{6.87}$^\dagger$ & \underline{2.96}$^\dagger$ & \underline{3.74}$^\dagger$ \\
\rowcolor{gray!20}$\Delta_{\mathrm{LETTER}}$ & +29.0\% & +34.1\% & +26.1\% & +29.0\% & +10.2\% & +11.2\% & +9.1\% & +9.9\% \\
 \midrule
Spectral & -- & -- & -- & -- & 4.17 & 6.35 & 2.76 & 3.45 \\
LETTER & 2.56 & 3.43 & 1.90 & 2.18 & 4.02 & 6.18 & 2.71 & 3.41 \\
TIGER & 2.34 & 3.11 & 1.72 & 1.97 & 3.69 & 5.84 & 2.44 & 3.13 \\
S$^2$GR & \underline{2.91} & \underline{3.95} & \underline{2.11} & \underline{2.45} &  3.68 & 5.74 & 2.41 & 3.07 \\
MMGRec & -- & -- & -- & -- & 3.17 & 5.08 & 2.06 & 2.67 \\
HSTU & 1.92 & 3.32 & 1.12 & 1.57 & 3.27 & 5.94 & 1.74 & 2.60 \\
SASRec & 1.03 & 1.71 & 0.64 & 0.86 & 1.68 & 2.96 & 1.03 & 1.44 \\
\midrule
& \multicolumn{4}{c|}{\textbf{MIND}} & \multicolumn{4}{c}{\textbf{Sports}} \\
\midrule
\textbf{RecDMoN} & 10.12 & 15.69 & \underline{6.71} & 8.50 & \textbf{2.89}$^\dagger$ & \textbf{4.72}$^\dagger$ & \textbf{1.83}$^\dagger$ & \textbf{2.41}$^\dagger$ \\
 \rowcolor{gray!20}$\Delta_{\mathrm{LETTER}}$ & +21.3\% & +18.0\% & +21.5\% & +19.6\% & +36.3\% & +38.0\% & +32.4\% & +34.2\% \\
 \textbf{RQ-GAE} & 8.33 & 13.49 & 5.55 & 7.21 & \underline{2.28}$^\dagger$ & \underline{3.75}$^\dagger$ & \underline{1.45} & \underline{1.92}$^\dagger$ \\
 \rowcolor{gray!20}$\Delta_{\mathrm{LETTER}}$ & -0.1\% & +1.5\% & +0.4\% & +1.3\% & +7.4\% & +9.6\% & +4.9\% & +6.8\% \\
 \midrule
 Spectral & \underline{10.24} & \underline{16.14} & 6.64 & \underline{8.54} & 1.70 & 2.94 & 1.05 & 1.45 \\
 LETTER & 8.34 & 13.29 & 5.53 & 7.11 & 2.12 & 3.42 & 1.38 & 1.80 \\
 TIGER & 6.32 & 9.53 & 4.16 & 5.19 & 2.12 & 3.46 & 1.33 & 1.76 \\
 S$^2$GR & 4.31 & 6.90 & 2.90 & 3.73 & 2.02 & 3.24 & 1.28 & 1.68 \\
 MMGRec & -- & -- & -- & -- & 1.62 & 2.73 & 1.02 & 1.38 \\
 HSTU & \textbf{12.16} & \textbf{18.32} & \textbf{7.98} & \textbf{9.96} & 2.02 & 3.41 & 1.07 & 1.52 \\
 SASRec & 8.07 & 12.96 & 5.30 & 6.87 & 1.21 & 2.00 & 0.79 & 1.04 \\
\midrule
& \multicolumn{4}{c|}{\textbf{Yelp}} & \multicolumn{4}{c}{\textbf{Toys}} \\
\midrule
\textbf{RecDMoN} & -- & -- & -- & -- & \textbf{4.81} & \textbf{8.16}$^\dagger$ & \textbf{2.94}$^\dagger$ & \textbf{4.02}$^\dagger$ \\
 \rowcolor{gray!20}$\Delta_{\mathrm{LETTER}}$ & -- & -- & -- & -- & +34.6\% & +52.0\% & +22.9\% & +35.3\% \\
  \textbf{RQ-GAE} & \underline{1.56} & \underline{2.68} & \underline{0.97} & \underline{1.33} & 4.22 & 6.51 & \underline{2.76}$^\dagger$ & \underline{3.50}$^\dagger$ \\
 \rowcolor{gray!20}$\Delta_{\mathrm{LETTER}}$ & +73.3\% & +83.5\% & +63.7\% & +72.2\% & +18.1\% & +21.3\% & +15.4\% & +17.9\% \\
 \midrule
 Spectral & \textbf{2.62} & \textbf{4.30} & \textbf{1.69} & \textbf{2.23} & 3.50 & 5.33 & 2.28 & 2.87 \\
 LETTER & 0.90 & 1.46 & 0.60 & 0.77 & 3.57 & 5.37 & 2.39 & 2.97 \\
 TIGER & 1.31 & 2.14 & 0.84 & 1.11 & 3.26 & 5.24 & 2.04 & 2.68 \\
 S$^2$GR & 1.49 & 2.54 & 0.94 & 1.28 & 3.33 & 5.40 & 2.10 & 2.77 \\
 MMGRec & -- & -- & -- & -- & 2.29 & 3.67 & 1.42 & 1.86 \\
 HSTU & 1.28 & 2.38 & 0.77 & 1.12 & \underline{4.35} & \underline{7.03} & 2.28 & 3.15 \\
 SASRec & 1.40 & 2.39 & 0.87 & 1.19 & 1.35 & 2.29 & 0.88 & 1.18 \\
\bottomrule
\end{tabular}%
\end{table*}

Table~\ref{tab:results_split_by_dataset} compares RecDMoN and RQ-GAE against atomic-ID and semantic-ID baselines.
All methods use the same TIGER (T5-based) generative recommendation backbone.
On the three small Amazon benchmarks (Beauty, Sports, and Toys), RecDMoN achieves the best results among all evaluated methods.
RQ-GAE ranks among the top two methods on Books, Beauty, Sports, and Yelp across all reported metrics.
These results suggest that collaborative graph structure provides useful signal for SID construction.
The signal is complementary to purely semantic or CF-embedding-based identifiers.
Since the generative recommender backbone is fixed, the observed gains are attributable to the SID construction strategy rather than to a stronger downstream recommender.

On Books, RecDMoN could not be evaluated because the current DMoN implementation exceeded GPU memory limits.
RQ-GAE scales to this setting and achieves the strongest result among the evaluated methods, reaching 4.60 H@10.
Compared with S2GR and the CF-aware LETTER baseline, this corresponds to relative gains of $+16.5\%$ and $+34.1\%$, respectively.

\paragraph{Discussion on Yelp and MIND}
The Yelp and MIND results provide a useful view of how GrIS behaves beyond the Amazon benchmarks.
On Yelp, RQ-GAE remains the second-best method across all reported metrics.
On MIND, HSTU achieves the strongest overall performance, while RecDMoN remains competitive.
This behaviour is plausible because MIND differs structurally from the Amazon datasets.
It has substantially longer user histories on average and a relatively compact item set.
This may favour strong sequential models over SID-based generative methods.
Overall, results suggest that the effectiveness of graph-aware SID construction is clear but certain nuances can be observed depending on dataset structure, training stability, and scalability of the graph-clustering stage.

\subsubsection{Impact of Graph Construction and Graph-Aware Design Choices}

We conduct the graph construction ablation on the Beauty dataset and test a windowed co-occurrence graph partition leaving other graph construction strategies such as random walk proximity or session-based co-occurrence as future work.
Given each user's time-ordered interaction sequence, we construct item--item graphs using forward windowed co-occurrence.
The adjacent-pair graph is a special case of this construction with window size \(w=1\), where each item is linked only to the next item in the sequence.
We evaluate windowed graphs with \(w \in \{2,3,5\}\), where each item is linked to the next \(w\) items.
For the \(w=5\) setting, we additionally test inverse-distance weighting, where pairs at distance \(k\) receive weight \(1/k\).
Edge weights are summed over all users and occurrences, and graphs are symmetrised before being passed to the graph-aware SID module.

\begin{figure*}
    \centering
    \includegraphics[width=\linewidth]{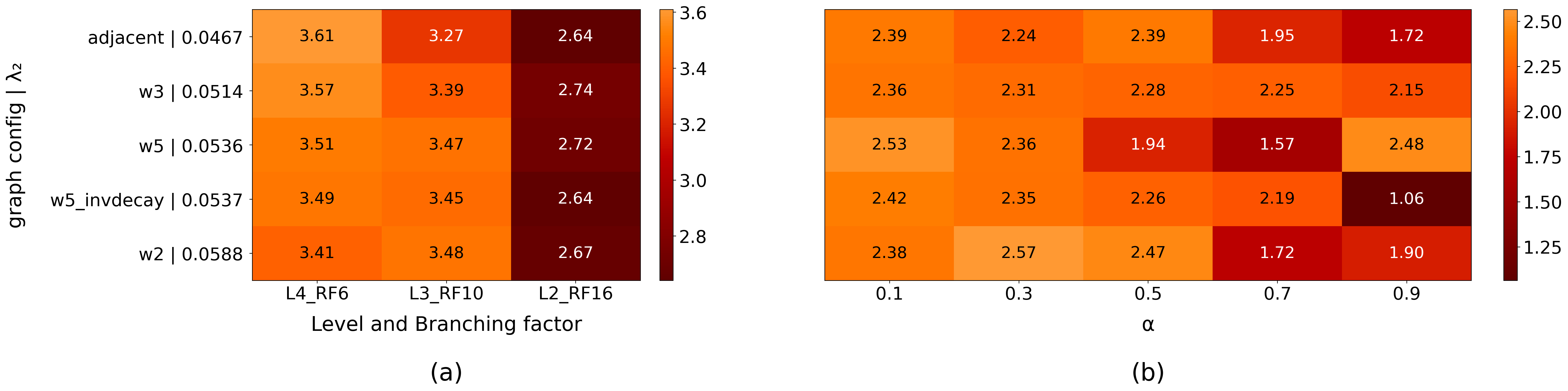}
    \caption{
        N@10 (\%) on Beauty across graph construction strategies for
        (a) Recursive DMoN with different hierarchy configurations and
        (b) RQ-GAE with different APPNP propagation coefficients.
    }
    \label{fig:beauty_graph_topology_heatmaps}
\end{figure*}

Table~\ref{tab:beauty_graph_topology_stats} summarises the graph variants used in the ablation.
The table reports the number of edges $|E|$, the average node degree computed on the unweighted graph, the average weighted degree after aggregating co-occurrence weights, and $\lambda_2$, the second-smallest eigenvalue of the normalised graph Laplacian.
This eigenvalue is commonly used as a spectral proxy for graph connectivity: larger values generally indicate stronger global connectivity.
We report the ablation as heatmaps in Figure~\ref{fig:beauty_graph_topology_heatmaps}, where rows correspond to graph variants and columns correspond to method-specific hyperparameters.
For RecDMoN, we vary the hierarchy depth and recursive branching factor using \(L4\_RF6\), \(L3\_RF10\), and \(L2\_RF16\).
For RQ-GAE, we vary the APPNP propagation coefficient \(\alpha \in \{0.1,0.3,0.5,0.7,0.9\}\), where larger \(\alpha\) preserves more of the initial semantic representation and reduces the relative contribution of neighbourhood propagation.
To keep the computational cost manageable, we reduce the number of RQ-GAE training iterations from \(20\text{k}\) to \(10\text{k}\) for this ablation.
Each heatmap cell reports the mean over three random seeds.
This setup allows us to study how graph construction interacts with the main graph-aware SID design choices.

Figure~\ref{fig:beauty_graph_topology_heatmaps}(a) shows the results for RecDMoN.
The strongest results are obtained with the deeper \(L4\_RF6\) configuration, suggesting that a more gradual coarse-to-fine hierarchy is beneficial for downstream generative recommendation.
Compared with shallower alternatives, this setting performs recursive clustering through more levels with a smaller branching factor at each level, which may better align the graph hierarchy with the prefix structure of semantic IDs.

Figure~\ref{fig:beauty_graph_topology_heatmaps}(b) shows the results for RQ-GAE.
The performance is generally stronger for smaller APPNP coefficients, indicating that neighbourhood propagation is important for constructing effective graph-aware IDs.
In contrast, very large \(\alpha\) values make the propagated representation closer to the original semantic embedding and lead to less stable downstream performance.
This suggests that graph-aware SID quality depends on retaining semantic information while still allowing sufficient collaborative signal to enter the quantised representation.

Overall, both heatmaps highlight the importance of method-specific graph-aware hyperparameters, especially hierarchy design for RecDMoN and propagation strength for RQ-GAE.
However, we do not observe a simple monotonic relationship between the reported graph statistics and downstream recommendation quality.
This indicates that the effect of graph construction is mediated by the SID generation method, and a more detailed analysis of graph connectivity and assignment quality remains an important direction for future work.

\begin{table}
  \centering
  \caption{Graph statistics for the Beauty graph construction ablation.}
  \label{tab:beauty_graph_topology_stats}
  \begin{tabular}{lrrrr}
    \toprule
    Graph variant & $|E|$ & Avg. deg. & Avg. w-deg. & $\lambda_2$ \\
    \midrule
    adjacent             & 111649 & 18.45 & 21.72 & 0.0467 \\
    w2           & 198768 & 32.85 & 39.74 & 0.0588 \\
    w3           & 265133 & 43.82 & 54.07 & 0.0514 \\
    w5           & 361874 & 59.81 & 75.88 & 0.0536 \\
    w5\_invdecay & 361874 & 59.81 & 40.46 & 0.0537 \\
    \bottomrule
  \end{tabular}
\end{table}

\subsubsection{RQ-GAE Components Ablation}

Table~\ref{tab:rq_gae_ablation} reports the contributions of graph-informed item representations and graph reconstruction regularisation to the effectiveness of RQ-GAE across three datasets. Compared with the base TIGER model (bottom row), introducing only the graph contrastive loss consistently improves performance across all three datasets, indicating that graph-structured supervision serves as a robust regulariser for semantic ID learning. In contrast, using only graph-informed inputs yields mixed results: it brings modest gains on Toys but underperforms the base model on Beauty and Sports, suggesting that directly replacing the original item semantic embedding with a non-parametric graph representation may introduce noise or weaken sequentially relevant semantics. The full model, which combines both components, achieves the best overall performance, ranking first on Beauty and Toys and remaining competitive on Sports. This shows that the two components are complementary: graph-informed inputs inject structural relational information into the encoder, while graph contrastive loss regularises the latent space to preserve graph proximity during quantisation.

\begin{table}
    \centering
    \caption{RQ-GAE components ablation. Graph Input refers to a non-parametric graph-informed item representation (APPNP) and Graph Loss refers to the graph-contrastive loss at each codebook level. Bold scores indicate leading results while underscore indicates second best results.}
    \label{tab:rq_gae_ablation}
    \resizebox{\columnwidth}{!}{%
    \begin{tabular}{cc|cccccc}
    \toprule
    \multirow{2}{*}{\makecell{Graph\\Input}} & \multirow{2}{*}{\makecell{Graph\\Loss}} & \multicolumn{2}{c}{Beauty} & \multicolumn{2}{c}{Sports} & \multicolumn{2}{c}{Toys} \\
    \cmidrule(lr){3-4}\cmidrule(lr){5-6}\cmidrule(lr){7-8}
     & & H@10 & N@10 & H@10 & N@10 & H@10 & N@10 \\
    \midrule
    \checkmark & \checkmark & \textbf{6.87} & \textbf{3.74} & \underline{1.92} & \underline{3.75} & \textbf{3.50} & \textbf{6.51}\\
     & \checkmark & \underline{6.82} & \underline{3.66} & \textbf{1.95} & \textbf{3.79} & \underline{3.30} & \underline{6.22} \\
    \checkmark &  & 5.66 & 3.10 & 1.49 & 2.95 & 2.78 & 5.38 \\
     & & 5.84 & 3.13 & 1.76 & 3.46 & 2.68 & 5.24\\
    \bottomrule
    \end{tabular}}
\end{table}

\subsubsection{Graph Clustering Analysis}

Table~\ref{tab:cluster_sizes} shows that the superiority of RecDMoN over spectral clustering is not limited to the raw number of clusters per level, but also extends to the balance of the induced hierarchy. While both methods produce similar numbers of top-level clusters, their cluster-size distributions differ substantially. Across Beauty, Sports, and Toys, RecDMoN exhibits dramatically lower Gini indices from $L_1$ to $L_4$, indicating that items are distributed much more evenly across clusters at the upper and intermediate levels. In contrast, spectral clustering yields consistently high Gini values throughout the hierarchy, suggesting that a small number of clusters dominate the partition while many others remain underutilised. This imbalance implies poorer usage of the semantic ID space and less informative hierarchical prefixes. RecDMoN also allocates more clusters to $L_4$ and far fewer to 
$L_5$ (deduplication codebook), indicating that it resolves more structure before the terminal level. Although both methods show higher inequality at $L_5$, this is expected at the finest granularity; importantly, RecDMoN maintains balanced partitions through the earlier levels where coarse-to-fine semantic organisation is most critical. These results suggest that RecDMoN constructs a more balanced and effective hierarchical item vocabulary, which likely contributes to its stronger recommendation performance.

\begin{table}
\centering
\caption{Per-level number of clusters and Gini index for Spectral Clustering and RecDMoN semantic ID mappings.}
\label{tab:cluster_sizes}
\small
\setlength{\tabcolsep}{4pt}
\begin{tabular}{llrrrrrr}
\toprule
Dataset & Method & $L_1$ & $L_2$ & $L_3$ & $L_4$ & $L_5$ & Total \\
\midrule
\multirow{4}{*}{Beauty} & Spectral & 6 & 36 & 206 & 865 & 863 & 1976 \\
& \textit{Gini} & 0.736 & 0.815 & 0.845 & 0.811 &  0.774 & \\
 & RecDMoN & 6 & 35 & 210 & 1245 & 33 & 1529 \\
 & \textit{Gini} & 0.099 & 0.159 & 0.194 & 0.269 & 0.632 & \\
\midrule
\multirow{4}{*}{Sports} & Spectral & 6 & 36 & 210 & 1087 & 661 & 2000 \\
& \textit{Gini} & 0.517 & 0.655 & 0.730 & 0.727 & 0.813 & \\
 & RecDMoN & 6 & 36 & 216 & 1274 & 73 & 1605 \\
& \textit{Gini} & 0.177 & 0.198 & 0.225 & 0.271 & 0.753 & \\
\midrule
\multirow{4}{*}{Toys} & Spectral & 6 & 36 & 206 & 1083 & 261 & 1592 \\
& \textit{Gini} & 0.508 & 0.651 & 0.695 & 0.685 & 0.781 & \\
 & RecDMoN & 6 & 36 & 212 & 1220 & 48 & 1522 \\
& \textit{Gini} & 0.058 & 0.161 & 0.215 & 0.291 & 0.735 & \\
\bottomrule
\end{tabular}
\end{table}

\subsubsection{Sentence Embedding Ablation}

Table~\ref{tab:dmon-gae-t5-qwen} shows that replacing \textit{sentence-t5-base} from the original TIGER model with \textit{Qwen3-Embedding-0.6B} consistently improves recommendation performance for both RecDMoN and RQ-GAE across Beauty, Sports, and Toys domains. The gains are especially pronounced for RecDMoN, suggesting that methods relying more directly on the raw semantic text space are more sensitive to the quality of the sentence embedder. In contrast, RQ-GAE exhibits smaller but still consistent improvements, indicating that graph-aware representation learning and graph contrastive regularisation partially mitigate weaknesses in the underlying text embedding space. A plausible explanation is that Qwen produces higher-quality item representations due to its stronger model capacity, larger context window size, more retrieval-oriented embedding training, and potentially better utilisation of long and attribute-rich product text. These properties likely yield a better-structured continuous semantic space, which in turn facilitates more informative discretisation into semantic IDs.

\begin{table}
\centering
\caption{Results for RecDMoN and RQ-GAE with \textit{sentence-t5-base} (T5) and\textit{ Qwen3-Embedding-0.6B} (Qwen) sentence embedders. Bold scores indicate leading results.}
\label{tab:dmon-gae-t5-qwen}
\resizebox{\columnwidth}{!}{%
\begin{tabular}{llcccccc}
\toprule
\multirow{2}{*}{Method} & \multirow{2}{*}{Embedder} & \multicolumn{2}{c}{\textbf{Beauty}} & \multicolumn{2}{c}{\textbf{Sports}} & \multicolumn{2}{c}{\textbf{Toys}} \\
\cmidrule(lr){3-4} \cmidrule(lr){5-6} \cmidrule(lr){7-8}
 &  & H@10 & N@10 & H@10 & N@10 & H@10 & N@10 \\
\midrule
RecDMoN & Qwen & \textbf{7.96} & \textbf{4.09} & \textbf{4.72} & \textbf{2.41} & \textbf{8.16} & \textbf{4.02} \\
RecDMoN & T5 & 7.24 & 3.71 & 4.47 & 2.31 & 7.29 & 3.65 \\
\midrule
RQ-GAE & Qwen & \textbf{6.87} & \textbf{3.74} & \textbf{3.75} & \textbf{1.92} & \textbf{6.51} & \textbf{3.50} \\
RQ-GAE & T5 & 6.85 & 3.72 & 3.57 & 1.85 & \textbf{6.51} & 3.43 \\
\bottomrule
\end{tabular}%
}
\end{table}

\subsubsection{Final Remarks on RecDMoN vs.\ RQ-GAE}
The results reveal a clear performance split across dataset scales.
RecDMoN outperforms all baselines on smaller datasets, suggesting that differentiable graph pooling is a powerful hierarchical partition mechanism when the full graph can be materialised.
However, in its current implementation RecDMoN requires storing the dense adjacency matrix, making it impractical for larger item catalogues; on those datasets we report RQ-GAE results only.
RQ-GAE proves competitive or dominant at scale, confirming that augmenting the RQ-VAE objective with a graph reconstruction term is an effective and scalable way to inject collaborative signal into the partition axis.

It is worth emphasising that RecDMoN and RQ-GAE are not merely two implementations of the same idea.
They represent qualitatively different families of hierarchical partitioning: 
RecDMoN performs a global, differentiable assignment over the graph via soft cluster memberships, whereas RQ-GAE inherits the greedy, quantisation-based commitment of the RQ family and augments it with a graph-aware reconstruction objective.

\section{Conclusions}\label{sec:conclusion}
We introduce GrIS, a framework that formalises semantic ID construction as 
hierarchical graph partition over a collaborative item graph.
By making explicit the two axes of graph construction and hierarchical 
partition, GrIS provides a principled design space for SID methods and a 
natural account of content-only approaches as empty-graph special cases.
Two instantiations, RecDMoN and RQ-GAE, demonstrate that different  
partition mechanisms can be accommodated within the framework while both 
improving over strong collaborative-filtering-aware baselines.

The framework opens concrete directions for future work on both axes.
On graph construction, replacing the co-occurrence weighted graph with a 
knowledge graph would allow external entity relationships and item attributes 
to inform the partition directly, enriching the collaborative signal with 
structured world knowledge. 
Beyond the local co-occurrence structures evaluated here, future work could investigate longer-range item relationships captured through session-level co-appearance and random-walk proximity.
On hierarchical partition, extending the partition target from individual 
items to item $n$-grams would allow sequence length to be compressed 
during tokenisation: mapping $n$-grams to $k$ codes increases sequence 
length by a factor of $k/n$ rather than $k$, a direct lever on computational cost, given that transformer architectures scale as 
$\Omega(L^2)$ in sequence length, making sequence compression a direct lever 
on computational cost.

One inherent limitation of any method that incorporates collaborative signal 
is temporal drift: as user behaviour evolves, the item graph changes, and 
previously derived partitions become stale. We leave a systematic study of incremental graph and partition updates to 
future work. Current implementation of RecDMoN limits its applicability to large item catalogues. Developing sparse, sampled, or mini-batch variants is therefore an important direction.

\section{GenAI Usage Disclosure}\label{sec:disclosure}
Generative AI tools, including ChatGPT, were used for non-substantive
assistance with language polishing, grammar correction, and writing
suggestions, as well as for assistance with writing
simple routine code. All AI-assisted text and code were
reviewed, edited where necessary, and verified or tested by the authors.
The research ideas, methodology, experimental design, analysis, and
conclusions were developed by the authors, who take full responsibility
for the work.


\bibliographystyle{ACM-Reference-Format}
\bibliography{bibliography}

\appendix

\section{SID Quality Metrics}

\begin{table*}
\caption{SID-space diagnostics and per-level codebook cardinalities on all six datasets. $L$ denotes the number of codebooks, while $\delta_{\mathrm{CF}}$ and $\delta_{\mathrm{Sem}}$ denote prefix-level cosine-similarity gaps computed using SASRec and \textit{Qwen3-Embedding-0.6B} embeddings, respectively.}

\label{tab:sid_analysis}
    \resizebox{\linewidth}{!}{%
\begin{tabular}{ll|rr|rrr|rrr|rrr|rrr|rrr}
\toprule
 & \multirow{2}{*}{Model} & \multirow{2}{*}{L} & \multirow{2}{*}{Collision Rate} & \multicolumn{3}{c|}{$|\mathcal{C}^l|$} & \multicolumn{3}{c|}{Gini} & \multicolumn{3}{c|}{Utilisation Ratio} & \multicolumn{3}{c|}{$\delta_{Sem}$} & \multicolumn{3}{c}{$\delta_{CF}$} \\
\cmidrule(lr){5-7} \cmidrule(lr){8-10} \cmidrule(lr){11-13} \cmidrule(lr){14-16} \cmidrule(lr){17-19}
 &  & &  & $l=0$ & $l=1$ & $l=2$ & $l=0$ & $l=1$ & $l=2$ & $l=0$ & $l=1$ & $l=2$ & $l<1$ & $l<2$ & $l<3$ & $l<1$ & $l<2$ & $l<3$ \\
\midrule
\multirow{7}{*}{\rotatebox[origin=c]{90}{Beauty}} & RecDMoN & 5 & 0 & 7 & 36 & 211 & 0.224 & 0.179 & 0.194 & - & - & - & 0.074 & 0.122 & 0.177 & -0.002 & -0.006 & 0.005 \\
 & RQ-GAE & 4 & 0 & 256 & 256 & 256 & 0.147 & 0.222 & 0.156 & 1 & 1 & 1 & 0.088 & 0.184 & 0.322 & -0.010 & 0.001 & -0.022 \\
 & LETTER & 4 & 0 & 44 & 256 & 256 & 0.504 & 0.178 & 0.192 & 0.172 & 1 & 1 & 0.050 & 0.118 & 0.220 & -0.005 & 0.012 & 0.018 \\
 & MMGRec & 5 & 0 & 256 & 78 & 136 & 0.224 & 0.918 & 0.860 & 1 & 0.305 & 0.531 & 0.074 & 0.085 & 0.100 & 0.010 & -0.002 & 0.001 \\
 & S$^2$GR & 4 & 0.001 & 44 & 250 & 244 & 0.297 & 0.197 & 0.187 & 0.172 & 0.977 & 0.953 & 0.163 & 0.333 & 0.481 & 0 & -0.007 & -0.001 \\
 & Spectral & 5 & 0 & 7 & 37 & 207 & 0.770 & 0.816 & 0.844 & - & - & - & 0.007 & 0.018 & 0.020 & -0.006 & 0.017 & 0.010 \\
 & TIGER & 4 & 0.002 & 46 & 256 & 256 & 0.294 & 0.179 & 0.165 & 0.180 & 1 & 1 & 0.170 & 0.353 & 0.513 & -0.014 & -0.006 & 0 \\
\midrule
\multirow{7}{*}{\rotatebox[origin=c]{90}{Sports}} & RecDMoN & 5 & 0 & 7 & 37 & 217 & 0.290 & 0.214 & 0.225 & - & - & - & 0.081 & 0.118 & 0.178 & 0.091 & 0.177 & 0.228 \\
 & RQ-GAE & 4 & 0 & 256 & 256 & 256 & 0.157 & 0.231 & 0.146 & 1 & 1 & 1 & 0.088 & 0.127 & 0.299 & 0.273 & 0.410 & 0.487 \\
 & LETTER & 4 & 0.003 & 68 & 256 & 256 & 0.434 & 0.172 & 0.193 & 0.266 & 1 & 1 & 0.072 & 0.094 & 0.290 & 0.281 & 0.562 & 0.505 \\
 & MMGRec & 5 & 0 & 256 & 93 & 93 & 0.180 & 0.917 & 0.865 & 1 & 0.363 & 0.363 & 0.071 & 0.092 & 0.097 & 0.210 & 0.235 & 0.259 \\
 & S$^2$GR & 4 & 0.001 & 27 & 256 & 256 & 0.311 & 0.188 & 0.160 & 0.105 & 1 & 1 & 0.148 & 0.321 & 0.490 & 0.145 & 0.213 & 0.272 \\
 & Spectral & 5 & 0 & 7 & 37 & 211 & 0.581 & 0.659 & 0.729 & - & - & - & 0.040 & 0.055 & 0.065 & 0.070 & 0.143 & 0.192 \\
 & TIGER & 4 & 0.005 & 256 & 256 & 256 & 0.417 & 0.181 & 0.179 & 1 & 1 & 1 & 0.224 & 0.445 & 0.547 & 0.184 & 0.244 & 0.299 \\
 \midrule
\multirow{7}{*}{\rotatebox[origin=c]{90}{Toys}} & RecDMoN & 5 & 0 & 7 & 37 & 213 & 0.188 & 0.179 & 0.214 & - & - & - & 0.051 & 0.126 & 0.188 & 0.059 & 0.122 & 0.210 \\
 & RQ-GAE & 4 & 0 & 256 & 256 & 256 & 0.144 & 0.178 & 0.137 & 1 & 1 & 1 & 0.096 & 0.232 & 0.363 & 0.304 & 0.549 & 0.686 \\
 & LETTER & 4 & 0.002 & 48 & 256 & 256 & 0.495 & 0.241 & 0.243 & 0.188 & 1 & 1 & 0.052 & 0.106 & 0.054 & 0.235 & 0.573 & 0.847 \\
 & MMGRec & 5 & 0 & 256 & 84 & 129 & 0.183 & 0.931 & 0.867 & 1 & 0.328 & 0.504 & 0.078 & 0.090 & 0.097 & 0.228 & 0.234 & 0.249 \\
 & S$^2$GR & 4 & 0 & 35 & 249 & 233 & 0.255 & 0.191 & 0.202 & 0.137 & 0.973 & 0.910 & 0.137 & 0.341 & 0.493 & 0.112 & 0.255 & 0.382 \\
 & Spectral & 5 & 0 & 7 & 37 & 207 & 0.572 & 0.655 & 0.694 & - & - & - & 0.014 & 0.021 & 0.030 & 0.039 & 0.093 & 0.141 \\
 & TIGER & 4 & 0.001 & 255 & 256 & 256 & 0.439 & 0.185 & 0.189 & 0.996 & 1 & 1 & 0.226 & 0.458 & 0.555 & 0.169 & 0.330 & 0.393 \\
\midrule
\multirow{4}{*}{\rotatebox[origin=c]{90}{Books}} & RQ-GAE & 4 & 0.002 & 256 & 256 & 256 & 0.288 & 0.337 & 0.281 & 1 & 1 & 1 & 0.116 & 0.212 & 0.301 & 0.130 & 0.152 & 0.167 \\
 & LETTER & 4 & 0.079 & 256 & 256 & 256 & 0.409 & 0.387 & 0.376 & 1 & 1 & 1 & 0.208 & 0.430 & 0.510 & 0.044 & 0.061 & 0.061 \\
 & S$^2$GR & 4 & 0.023 & 88 & 256 & 256 & 0.275 & 0.154 & 0.154 & 0.344 & 1 & 1 & 0.246 & 0.481 & 0.591 & 0.025 & 0.068 & 0.088 \\
 & TIGER & 4 & 0.079 & 92 & 256 & 256 & 0.296 & 0.149 & 0.162 & 0.359 & 1 & 1 & 0.273 & 0.499 & 0.599 & 0.017 & 0.070 & 0.078 \\
 
\midrule
 \multirow{6}{*}{\rotatebox[origin=c]{90}{MIND}} & RecDMoN & 5 & 0 & 7 & 37 & 217 & 0.180 & 0.083 & 0.094 & - & - & - & 0.084 & 0.127 & 0.164 & 0.053 & 0.105 & 0.174 \\
 & RQ-GAE & 4 & 0 & 256 & 255 & 255 & 0.291 & 0.599 & 0.429 & 1 & 0.996 & 0.996 & 0.022 & 0.055 & 0.077 & 0.371 & 0.378 & 0.416 \\
 & LETTER & 4 & 0.005 & 167 & 256 & 256 & 0.225 & 0.215 & 0.222 & 0.652 & 1 & 1 & 0.022 & 0.078 & 0.452 & 0.241 & 0.475 & 0.376 \\
 & S$^2$GR & 4 & 0 & 37 & 256 & 256 & 0.217 & 0.153 & 0.125 & 0.145 & 1 & 1 & 0.170 & 0.303 & 0.448 & 0.103 & 0.171 & 0.206 \\
 & Spectral & 5 & 0 & 7 & 37 & 217 & 0.191 & 0.143 & 0.253 & - & - & - & 0.006 & 0.001 & 0.013 & 0.158 & 0.377 & 0.424 \\
 & TIGER & 4 & 0.006 & 207 & 256 & 256 & 0.379 & 0.160 & 0.141 & 0.809 & 1 & 1 & 0.223 & 0.387 & 0.521 & 0.138 & 0.195 & 0.219 \\
\midrule
\multirow{4}{*}{\rotatebox[origin=c]{90}{Yelp}} & RQ-GAE & 4 & 0.001 & 256 & 256 & 256 & 0.284 & 0.368 & 0.291 & 1 & 1 & 1 & 0.133 & 0.166 & 0.182 & 0.098 & 0.131 & 0.140 \\
 & LETTER & 4 & 0.002 & 256 & 256 & 256 & 0.131 & 0.089 & 0.093 & 1 & 1 & 1 & 0.017 & 0.069 & 0.191 & 0.017 & 0.023 & 0.044 \\
 & S$^2$GR & 4 & 0.001 & 256 & 256 & 256 & 0.473 & 0.255 & 0.204 & 1 & 1 & 1 & 0.174 & 0.267 & 0.337 & 0.085 & 0.097 & 0.105 \\
 & Spectral & 5 & 0 & 7 & 37 & 217 & 0.605 & 0.620 & 0.646 & - & - & - & 0.065 & 0.115 & 0.122 & -0.014 & 0.070 & 0.090 \\
 & TIGER & 4 & 0.002 & 256 & 256 & 256 & 0.528 & 0.230 & 0.263 & 1 & 1 & 1 & 0.182 & 0.268 & 0.345 & 0.080 & 0.097 & 0.104 \\
\bottomrule
\end{tabular}}
\end{table*}

Beyond downstream accuracy, Table~\ref{tab:sid_analysis} evaluates SID structure across all six datasets. For the first three codebook levels, we report cardinality, utilisation, and usage Gini, where higher utilisation and lower Gini indicate better coverage and balance. We also report the collision rate of complete SIDs, where lower is better.

Hierarchical coherence is measured by treating SID prefixes as clusters and computing $\delta=\mu_{\mathrm{intra}}-\mu_{\mathrm{inter}}$, where each mean cosine similarity is estimated from 1,000 sampled item pairs. $\delta_{\mathrm{CF}}$ uses the SASRec embeddings employed in LETTER training, whereas $\delta_{\mathrm{Sem}}$ uses \textit{Qwen3-Embedding-0.6B} embeddings. A larger positive $\delta$ indicates that items sharing a prefix are more similar than items assigned to different prefixes. Together, these metrics characterise the uniqueness, coverage, balance, and hierarchical organisation of the generated SIDs beyond their downstream recommendation accuracy.

CF-aware methods generally improve $\delta_{\mathrm{CF}}$ over TIGER at the cost of $\delta_{\mathrm{Sem}}$. Graph-aware methods preserve more semantic coherence, with RQ-GAE providing the strongest overall balance between CF and semantic structure while maintaining balanced codeword usage, which could explain its good overall downstream performance.

RecDMoN and spectral clustering use fewer than 10 first-level codewords, making their $\delta$ values less comparable with those of finer-grained methods. Between them, RecDMoN achieves higher $\delta_{\mathrm{Sem}}$ and $\delta_{\mathrm{CF}}$, consistent with its trainable graph-pooling architecture, which combines semantic node representations with collaborative graph structure and optimises modularity.

\end{document}